\documentclass[11pt,a4paper]{article}
\usepackage[hyperref]{emnlp2020}
\usepackage{times}
\usepackage{latexsym}
\usepackage{amsmath}
\usepackage{changepage}
\usepackage{graphicx}

\usepackage{float}
\usepackage{cleveref}
\usepackage{booktabs}
\usepackage{multirow}
\usepackage{makecell}
\usepackage{algorithm,algorithmic}
\usepackage{mathtools}
\usepackage[normalem]{ulem}
\usepackage{color,soul}
\usepackage{subfig}

\definecolor{celadon}{rgb}{0.67, 0.88, 0.69}
\definecolor{salmonpink}{rgb}{1.0, 0.57, 0.64}
\definecolor{lightpink}{rgb}{1.0, 0.71, 0.76}
\definecolor{beaublue}{rgb}{0.74, 0.83, 0.9}
\definecolor{indianyellow}{rgb}{0.89, 0.66, 0.34}
\definecolor{lightapricot}{rgb}{0.99, 0.84, 0.69}
\newcommand{\hlc}[2][yellow]{{\sethlcolor{#1}\hl{#2}} }
\newcommand{\hlp}[2][beaublue]{{\sethlcolor{#1}\hl{#2}} }
\newcommand{\hln}[2][lightapricot]{{\sethlcolor{#1}\hl{#2}} }
\definecolor{lemonchiffon}{rgb}{1.0, 0.98, 0.8}

\usepackage{microtype}

\aclfinalcopy 

\def\showcomments{1}
\if\showcomments1
\newcommand{\dd}[1]{\textcolor{red}{[  #1 -- Danish]}\typeout{#1}}
\newcommand{\bd}[1]{\textcolor{orange}{[  #1 -- Bhuwan]}\typeout{#1}}
\newcommand{\gn}[1]{\textcolor{purple}{[  #1 -- GN]}\typeout{#1}}

\newcommand{\zl}[1]{\textcolor{blue}{[  #1 -- Zack]}\typeout{#1}}
\else
\newcommand{\dd}[1]{}
\newcommand{\bd}[1]{}
\newcommand{\gn}[1]{}
\newcommand{\zl}[1]{}
\fi 

\title{Weakly- and Semi-supervised Evidence Extraction}

\author{Danish Pruthi, Bhuwan Dhingra, Graham Neubig, Zachary C. Lipton \\
Carnegie Mellon University \\ Pittsburgh, USA \\
\texttt{\{ddanish, bdhingra, gneubig, zlipton\}@cs.cmu.edu}
}

\begin{document}
\maketitle
\begin{abstract}
For many prediction tasks, stakeholders desire 
not only predictions but also supporting evidence 
that a human can use to verify its correctness.
However, in practice, additional annotations marking supporting evidence
may only be available for a minority of training examples (if available at all). 
In this paper, we propose new methods to combine 
few evidence annotations (strong semi-supervision) 
with abundant document-level labels (weak supervision) 
for the task of evidence extraction. 
Evaluating on two classification tasks 
that feature evidence annotations,
we find that our methods outperform 
baselines adapted from the interpretability literature to our task.
Our approach yields substantial gains 
with as few as hundred evidence annotations.\footnote{
Code and datasets to reproduce our work are available 
at \url{https://github.com/danishpruthi/evidence-extraction}.}
\end{abstract}

\section{Introduction}
\label{sec:intro}

\begin{table*}
\small
\centering
\begin{tabular}{@{}lcc@{}}
\toprule
 & \textbf{Explanations}                    & \textbf{Evidence}                                                                                            \\ \midrule
\textbf{Objective} & \begin{tabular}[c]{@{}c@{}} Elucidate ``the reasons behind predictions''. \end{tabular} & \begin{tabular}[c]{@{}c@{}} Enable users to quickly verify the predictions. \end{tabular}     \\ \midrule
\begin{tabular}[c]{@{}c@{}} \textbf{Evaluation} \end{tabular} & \begin{tabular}[c]{@{}c@{}}   Explanations are specific to the model. \\ No ground truth explanations to compare  against.\end{tabular}    & \begin{tabular}[c]{@{}c@{}} Evidence is a characteristic of the task. \\ Can be compared against human-labeled evidence. \end{tabular} \\ \midrule
\begin{tabular}[c]{@{}c@{}} \textbf{Example} \end{tabular} & \begin{tabular}[c]{@{}c@{}}   A \hlc{horror} movie that lacks cohesion.\end{tabular}    & \begin{tabular}[c]{@{}c@{}} A horror movie that \hlc{lacks cohesion.} \end{tabular} \\ \bottomrule
\end{tabular}
\caption{Distinguishing local explanations from evidence snippets. In the illustrative example, the token horror is predictive of the negative sentiment as horror movies tend to get poorer ratings than movies from other genres~\cite{kaushik2019learning}, however, no expert would mark it to be the evidence justifying the negative review.} 
\label{tbl:differences}
\end{table*}

Despite the success of deep learning for countless prediction tasks,
practitioners often desire that these models not only be accurate
but also provide \emph{interpretations} or \emph{explanations}
\cite{caruana2015intelligible, weld2019challenge}.
Unfortunately, these terms lack precise meaning,
and across papers, such explanations are purported
to address a spectrum of desiderata so vast
that it seems unlikely that any one method 
could address them all \citep{lipton2018mythos}.  
In both computer vision \citep{simonyan2013deep, ribeiro2016should} 
and natural language processing \citep{lei2016rationalizing,ribeiro2016should, lehman2019inferring}, 
proposed explanation methods often take the form of 
highlighting salient features of the input.
These so-called \emph{local explanations} are intended 
to highlight features that elucidate 
``the reasons behind predictions''~\cite{ribeiro2016should}.
However, this characterization of the problem remains under-specified.

In this paper, we instead focus on supplementing predictions with \emph{evidence}, 
which we define as information that gives users the ability 
to quickly verify the correctness of predictions. 
Fortunately, for many problems, a localized portion of the input
can be sufficient for a human to validate the predicted label.
When classifying large images, a small patch containing a hamster
may be sufficient to verify the applicability of the ``hamster'' label.  
Similarly, in a long clinical note 
(as are common in electronic medical records), 
a single sentence may suffice to confirm a predicted diagnosis. 
This ability to verify results engenders trust among users, 
and increases adoption of the machine learning systems~\cite{dzindolet2003role, herlocker2000explaining}.
In Table~\ref{tbl:differences}, we outline the characteristic differences 
between local explanations and evidence.

Thus motivated, we cast our problem as learning to extract evidence
using both strong (semi-) and weak (full) supervision. 
The former takes the form of explicit, but scarce, 
manual annotations of evidence segments,
whereas the latter is provided by documents and their class labels
which we assume are relatively abundant.\footnote{While the task formulation is broadly applicable, we limit to text classification tasks for the scope of this work.}
Note that absent any strong supervision, 
the task may be fundamentally underspecified.
Features can be predictive of the label (due to confounding)
absent a direct semantic connection to the label.
For instance, in the IMDb movie review dataset 
the token ``horror'' is predictive of negative sentiment 
because horror movies tend to receive poorer ratings 
than movies from other genres~\cite{kaushik2019learning}.
However, no expert would mark it to be the evidence justifying the negative review.
In the extreme case where evidence annotations are available for all examples,
our task reduces to a standard multitask learning problem.
In the opposite extreme, where only weak supervision is available, 
we find ourselves back in the under-specified realm 
addressed by local explanations. 
While predictive tokens might be extracted using only weak supervision,
evidence extraction requires some amount of strong supervision.

Drawing inspiration from ~\citet{zaidan2008modeling}, 
who study the reverse problem---how to leverage marked evidence spans 
to improve classification performance,
we design an approach for optimizing the joint likelihood of class
labels and evidence spans, given the input examples. 
We factorize our objective such that 
we first \emph{classify, and then extract} the evidence.
For classification, we use 
BERT~\cite{devlin2018bert}.
The extraction task (a sequence tagging problem)
is modeled using a linear-chain CRF~\cite{lafferty2001conditional}.  
The CRF uses representations and attention scores 
from BERT as emission features, 
allowing the two tasks (classification and extraction) 
to benefit from shared parameters.
Further, the evidence extraction module is 
conditioned on the class label, 
enabling the CRF to output different evidence spans 
tailored to each class label.
This is illustrated in Table~\ref{tbl:qual_example}. 

For baselines, we repurpose 
input attribution methods 
from the interpretability literature.
Many approaches in this category
first \emph{extract, and then classify}~\cite{lei2016rationalizing,lehman2019inferring,jain2020learning,paranjape2020information}. 
Across two text sequence classification 
and evidence extraction tasks,
we find our methods to outperform 
baselines. Encouragingly, 
we observe gains by using our approach 
with as few as $100$ evidence annotations.

\begin{table*}
    \small
    \centering
    \begin{tabular}{@{}c@{}}
    \toprule
     \textbf{Movie Review}                                                                                             \\ \toprule
     \begin{tabular}[l]{@{}l@{}} I don't know what movie the critics saw, but it wasn't this one. The popular consensus among newspaper \\ critics was that \hln{\textbf{this movie is unfunny and dreadfully boring}}. In my personal opinion, they couldn't be more wrong.\\ If you were expecting Airplane! - like laughs and Agatha Christie - intense mystery, then yes, this movie would \\ be a disappointment. However, if you're just looking for \hlp{\textbf{an enjoyable movie and a good time}}, this is \hlp{\textbf{one}} to see ... \end{tabular} \\ \midrule
    \begin{tabular}[l]{@{}l@{}} Lean, mean, escapist thrillers are a tough product to come by. \hln{\textbf{Most are unnecessarily complicated}}, and others \\ have no sense of expediency--the thrill-ride effect gets lost in the \hln{\textbf{cumbersome}} plot. Perhaps the ultimate escapist\\ thriller was the fugitive, which featured none of the flash-bang effects of today's market but rather a bread-and-butter, \\ \hlp{\textbf{textbook example of what a clever script and good direction is all about.}} ... \end{tabular}\\ \midrule 
    \end{tabular}
    \caption{Non cherry-picked evidence extractions from our approach. We condition our extraction model on both the \hlp{\textbf{positive}} and the \hln{\textbf{negative}} label. Our approach is able to tailor the extractions as per the conditioned label.}
    \label{tbl:qual_example}
\end{table*}

\section{Related Work}
\label{sec:related_work}

We briefly discuss methods from the interpretability literature 
that aim to identify salient features of the input.  
\citet{lei2016rationalizing} propose an approach 
wherein a generator first extracts a subset of the text from the original input,
which is subsequently fed to an encoder that classifies the input
conditioned only on the extracted subset. 
The generator and encoder are trained end-to-end 
via REINFORCE-style optimization~\cite{williams1992simple}. 
However, follow-up work discovered the end-to-end training 
to be quite unstable with high variance in results~\cite{bastings2019interpretable,paranjape2020information}. 
Consequently, other approaches 
adopted the core idea of \emph{extract, and then classify}
in different forms: 
\citet{lehman2019inferring} decouple 
the extraction and prediction modules 
and train them individually with intermediate supervision;
\citet{jain2020learning} use  
heuristics, like attention scores, for extraction; 
and lastly, \citet{paranjape2020information}
extract subsets that have high mutual information with the output
variable and low mutual information with the input variable.

\section{Extracting Evidence}
\label{sec:methods}

Formally, let the training data consist of $n$ points 
$\{(\mathbf{x}_1, y_1)... (\mathbf{x}_n, y_n)\}$,
where  $\mathbf{x}_i$ is a document
and $y_i$ is the associated label. 
We assume that for $m$ points ($m \ll n$),
we also have evidence annotations $\mathbf{e}_i$, 
a binary vector such that $e_{ij} = 1$ 
if token $x_{ij}$ is a part of the evidence, and $0$ otherwise.
The conditional likelihood of the output labels and evidence,
given the documents, can be written as:   
\begin{equation*}
L = \prod_{i=1}^n p(y_i, \mathbf{e}_i | \mathbf{x}_i)
\end{equation*}
We can factorize this likelihood in two ways. First,
\begin{align*}
L &= \prod_{i=1}^n p(\mathbf{e}_i|\mathbf{x}_i)\ p(y_i | \mathbf{x}_i, \mathbf{e}_i)  \\
&= \ \prod_{i=1}^n \underbrace{\ p(\mathbf{e}_i|\mathbf{x}_i)\ }_\text{extract} \ \underbrace{p(y_i | \mathbf{e}_i)}_\text{classify} \\
& (\text{assuming } \  y_i \perp \mathbf{x}_i | \mathbf{e}_i)
\end{align*}
This corresponds to the \emph{extract, then classify} approach. 
Since both components of this likelihood function require extractions, 
supervised methods can only leverage 
$m$ (out of $n$) training examples~\cite{lehman2019inferring}. 
Unsupervised or semi-supervised extraction methods 
can still use all the document–level labels during 
training~\cite{jain2020learning,paranjape2020information}.
Alternatively, we can factorize the likelihood as follows:
\begin{align*}
L &= \prod_{i=1}^n \underbrace{\ p(y_i | \mathbf{x}_i)\ }_\text{classify} \  \underbrace{p(\mathbf{e}_i | y_i, \mathbf{x}_i)}_\text{extract}
\end{align*}
The \emph{classify, then extract}
approach is amenable to weakly supervised learning, 
as we can optimize the classification objective for all $n$ examples
and the extraction objective for $m$ examples.

We use BERT~\cite{devlin2018bert} 
to model $p_\theta(y|\mathbf{x})$ 
and a linear-chain CRF~\cite{lafferty2001conditional} 
to model $p_\phi(\mathbf{e}|\mathbf{x}, y; \theta)$, where
\begin{equation*}
    p_\phi(\mathbf{e} | y,\mathbf{x}) = \frac{1}{Z} \exp\left\{\sum_{k=1}^{K}\phi_k f_k(e_t, e_{t-1}, x_t, y)\right\}
\end{equation*}
Here $t$ indexes the input sequence, 
and $Z$ is a normalization factor. 
Function $f(\cdot)$ extracts $K$ features 
including both emission and transition features,
and $\phi$ are the corresponding weights.
The transition weights allow the CRF 
to model contiguity in the evidence tokens. 

We examine two types of emission features 
for a given token $x_t$ in the input $x$, including
(1) \textbf{BERT features} ($f_{\text{BERT}}(x)_{t}$)
where we encode the entire input sequence, 
and use the representation corresponding to token $x_t$;\footnote{Note that we share the BERT representations 
between the classification and extraction modules.} and 
(2) \textbf{attention features}
where we use the last layer attention values 
from different heads of the \texttt{[CLS]} token 
to the given token $x_t$.  
These features tie the classification and extraction architectures. 

The \textit{classify, then extract} approach also allows 
conditioning the evidence extraction model 
on the (predicted or oracle) \textbf{label} of the text document. 
For binary classification, one way to achieve this 
is to transform the existing emission features $f$ 
to new features $f^{'}$ in the following manner: 
\[
\small
f^{'}_{2k}(e_t, e_{t-1}, x_t, y)=\begin{cases}
    f_{k}(e_t, e_{t-1}, x_t) & \text{if}\ y = 0 \\
          0 & \text{if}\ y = 1
          \end{cases}
\]
\[
\small
f^{'}_{2k+1}(e_t, e_{t-1}, x_t, y)=\begin{cases}
     0 & \text{if}\ y = 0 \\
     f_{k}(e_t, e_{t-1}, x_t) & \text{if}\ y = 1
          \end{cases}
\]
This transformation allows us to use
even indexed emission weights $(\phi_{2k})$ for the first class, 
and odd indexed emission weights $(\phi_{2k+1})$ for the second class. 
Similar transformations can be easily constructed 
for multi-class classification problems.
During inference, we use the predicted label $\hat{y}$ instead of the true label $y$.
Using this formulation, emission features (and their corresponding weights) 
capture the association of each word 
with the extraction label (evidence or not) 
\emph{and} the classification label.
For instance, for binary sentiment analysis of movie reviews,
the token ``brilliant'' is highly associated with the positive class, 
and if the review is (marked/predicted to be) positive, 
then the chances to select it as a part of the evidence increase. 
Inversely, if ``brilliant'' occurs in a negative review, 
the chances of selecting it decrease.

By conditioning the extraction models on the classification label,
one can find supporting evidence tailored for each class
(as one can see in Table~\ref{tbl:qual_example}). 
This can be especially useful when the input examples exhibit
characteristics of multiple classes, or when
classification models are less certain about their predictions. 
In such cases, examining the extractions for 
each class could help validate the model behaviour.

\paragraph{Implementation Details}
We train both the classification and extraction modules simultaneously. 
For evidence extraction, the emission features of the CRF 
include BERT representations or its attention values
(depending upon the experiment). 
The same BERT model is also used for classification; 
thus, the two tasks share the BERT parameters. 
We use the transformers library by Hugging Face~\cite{Wolf2019HuggingFacesTS}, 
and default optimization parameters for finetuning BERT.  

\section{Results and Discussion}
\label{sec:results}
\begin{table*}[ht]
    \centering
    \small
    \begin{tabular}{@{}lccccccccc@{}}
    \toprule
    \multirow{3}{*}{Approach}                                                                             & \multicolumn{3}{c}{Sentiment Analysis}   & & & \multicolumn{3}{c}{Propaganda Detection}  & \\  \cline{2-4} \cline{7-9}
                                                                                &  \multicolumn{1}{c}{Prediction} &  & \multicolumn{1}{c}{Extraction} & & & \multicolumn{1}{c}{Prediction} &  & \multicolumn{1}{c}{Extraction}  \\ 
                                                                                                        & (Accuracy)                            &  & (F1 score)           & &  & (F1 score)                            &  & (F1 score)                   \\ \toprule
    \begin{tabular}[c]{@{}c@{}}Pipeline approach$^\diamond$~\cite{lehman2019inferring} \end{tabular}                          & 76.9                                &  & 14.0 & & & --- &  & ---                             \\ 
    \begin{tabular}[c]{@{}c@{}}Information Bottleneck (IB)$^{\dagger}$$^\diamond$~\cite{paranjape2020information} \end{tabular}                          & 82.4                                &  & 12.3 & & & --- &  & ---                             \\ 
    \begin{tabular}[c]{@{}c@{}}IB (semi-supervised, 25\%)$^\diamond$ ~\cite{paranjape2020information} \end{tabular}                          & 85.4                                 & & 18.1 & & & --- &  & ---                             \\ 
    \begin{tabular}[c]{@{}c@{}}Top-k attention$^\dagger$~\cite{jain2020learning}\end{tabular}                            & 93.1                           &  & 27.7 & & & 65.8 &  & 27.4 \\ 
    \begin{tabular}[c]{@{}c@{}}Supervised attention~\cite{zhong2019fine} \end{tabular}                                                                                & 93.2                           &  & 43.1 & & & 67.1 &  & 34.2 \\ \midrule
    Our Methods  \\
        \midrule
    \begin{tabular}[c]{@{}c@{}} Classify only (BERT) \end{tabular}                                                                                        & 93.1                           &  & --- & & & 65.8 &  & ---                  \\ 
    \begin{tabular}[c]{@{}c@{}} Extract only (BERT-CRF) 
    \end{tabular}                                                                                     &        ---                        &  & 42.6     & & & --- &  & 39.1                 \\ 
    \begin{tabular}[c]{@{}c@{}} Classify \& Extract (BERT's Attention-CRF)\end{tabular}        & 93.1                           &  & 45.2          & & & 65.8 &  & 41.0            \\ 
    \begin{tabular}[c]{@{}c@{}} Classify \& Extract (BERT-CRF)\end{tabular}              & 93.3                           &  & 45.4         & & & 64.1 &  &  \textbf{41.5}           \\ \vspace{0.75mm}
    \begin{tabular}[c]{@{}c@{}} Classify \& Extract (BERT-CRF) w/ predicted labels \end{tabular}               & 93.2                           &  & \textbf{46.3}    & & & 64.9 &  & 41.2                        \\ \midrule
    \begin{tabular}[c]{@{}c@{}} Classify \& Extract (BERT-CRF) w/ oracle labels \end{tabular}               & 93.3                           &  & 46.8    & & & 64.9 &  & 45.0                        \\ \bottomrule
    \end{tabular}
    \caption{Evaluating different methods on two classification tasks that feature evidence annotations. The last row is an upper bound assuming access to the oracle label for conditioning. $\dagger$ denotes unsupervised approaches, and $\diamond$ indicates sentence-level extraction methods, which can not be applied to the propaganda detection task as the input is only a single sentence. All the values are averaged across $5$ seeds. }
    \label{tab:results}
    \vspace{-0.1in}
\end{table*}

\vspace{5px}
\noindent \textbf{Baselines \quad} 
We use several approaches that attempt to rationalize predictions 
as baselines for the evidence extraction task. 
These include:
(i) the Pipeline approach~\cite{lehman2019inferring}, 
wherein the extraction and classification modules are pipe-lined,
with each individually trained with supervision; 
(ii) the Information Bottleneck approach~\cite{paranjape2020information},
which extracts sentences from the input such that
they have maximal mutual information (MI) with the output label, 
and minimal MI with the original input;\footnote{There 
exist trivial solutions to the Information Bottleneck objective 
when subset granularity is tokens instead of sentences.
One such solution is when the extraction model 
extracts ``.'' for the positive class and ``,'' for the negative class.}
(iii) the FRESH approach~\cite{jain2020learning}, 
which extracts the top-$k$ tokens with the highest attention scores
(value of $k$ is set to match the fraction of evidence tokens in the development set);\footnote{Interestingly,
\citet{jain2020learning} find this simple thresholding approach to be better than other 
end-to-end approaches~\cite{bastings2019interpretable,lei2016rationalizing}} 
and (iv) Supervised attention,
where attention is supervised to be uniformly high 
for tokens marked as evidence, and low otherwise~\cite{zhong2019fine}.

\vspace{5px}
\noindent \textbf{Setup \quad} We evaluate 
the different evidence extraction approaches 
on two text classification tasks: 
analyzing sentiment of movie reviews~\cite{pang2002thumbs}, 
and detecting propaganda techniques 
in news articles~\cite{EMNLP19DaSanMartino}.  
For the \textbf{sentiment analysis} task, 
we use the IMDb movie reviews dataset 
collected by ~\citet{maas-EtAl:2011:ACL-HLT2011},
comprising $25$K movie reviews available for training, 
and $25$K for development and testing. 
The dataset has disjoint sets of movies for training and testing.
Additionally, we use $1.8$K movie reviews 
with marked evidence spans collected by~\citet{zaidan2007using}. 
Of these $1.8$K spans, we use $1.2$K for training
and $300$ each for development and testing.
Note that here less than $5\%$ of all the movie reviews 
are annotated for evidence, 
and the reviews are consistently long
(with more than $600$ words on an average),
thus necessitating evidence to quickly verify the predictions.

For the task of \textbf{propaganda detection} in news articles, 
we use the binary sentence-level labels (propaganda or not), 
and token-level markings that support these labels.
Similar to the sentiment dataset,
we use token-level evidence markings for $5\%$ of all the sentences.  
The total number of sentences in train, dev, 
and test sets are $10.8$K, $1.7$K, $4$K respectively. 
Sentences without any propaganda content have no token-level markings.

\vspace{5px}
\noindent \textbf{Results \quad}
We evaluate the predictions and their supporting evidences from different models. 
We compute the micro-averaged token-wise F1 score for the extraction task.
From Table~\ref{tab:results}, we can clearly see that our approach 
outperforms other baseline methods on both the extraction tasks. 
The pipeline approach~\cite{lehman2019inferring}
is unable to leverage a large pool of classification labels.
Additionally, the pipeline and the Information Bottleneck approaches
extract evidence at a sentence level, 
whereas the evidence markings are at a token level, 
which further explains their low scores. 
Further, the top-$k$ attention baseline 
achieves a reasonable F1 score of $27.7$
on the extraction task for sentiment analysis task
and $27.4$ on the propaganda detection task,
without any supervision. 
This result corroborates the findings of~\citet{jain2020learning},
who demonstrated attention scores to be good heuristics for extraction.
Supervising attention with labeled extractions improves 
extraction score on both  tasks, 
which is inline with results in~\citet{zhong2019fine}.

In our approach, the extraction model
benefits from classification labels
because of two factors: 
(i) sharing parameters between extraction and classification; 
and (ii) conditioning on the predicted $\hat{y}$ for extraction.
These benefits are substantiated by comparing
the extract only (BERT-CRF) approach with the
classify \& extract (BERT-CRF) method.
The latter approach leads to improvements of $2.8$ and $2.4$ points 
for sentiment analysis and propaganda detection tasks, respectively.
Conditioning on the predicted label
improves the extractions by $0.9$ points on the sentiment analysis task.
For propaganda detection, we don't see an immediate benefit 
because many predicted labels are misclassified.
However, upon using oracle labels, 
the extraction performance improves by $3.5$ points.

\begin{figure}
\centering
    \subfloat[Sentiment Analysis]{{\includegraphics[width=4.25cm]{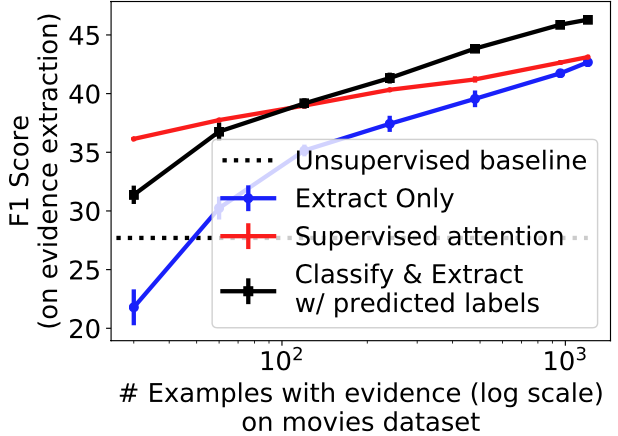}}}
    \subfloat[Propaganda detection]{{\includegraphics[width=3.66cm]{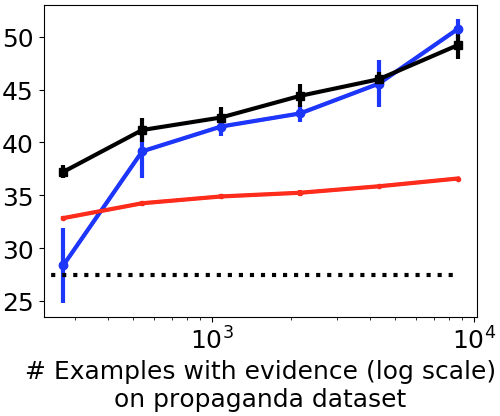}}}
\caption{Mean and standard error of extraction scores with increasing amounts of evidence annotations.
}
\vspace{-12px}
\label{fig:increasing_data}
\end{figure}

When we lower the number of evidence annotations available during training,
we discover (unsurprisingly) that the extraction performance 
degrades (Figure~\ref{fig:increasing_data}). 
For sentiment analysis, with less than $100$ annotations,
supervised attention performs the best, 
as no new parameters need to be trained.
However, with over $100$ training instances,
classify \& extract model outperforms this baseline
and is significantly better than the best unsupervised baseline.
For propaganda detection, our approaches perform the best.
As expected, the performance gap between 
extract only and classify \& extract 
approach decreases with increase in available annotations.

\section{Conclusion}
\label{sec:conclusion}

Despite the growing interest in producing \emph{local explanations}
in an attempt to elucidate ``the reasons behind predictions'', 
the underlying questions motivating this research remain ill-specified.
The problem is aggravated by a lack of standardized protocols 
to evaluate the quality of these explanations. 
In contrast, our work addresses the comparatively concrete problem
of supplementing predictions with \emph{evidence} 
that stakeholders can use to verify 
the correctness of the predictions.
Moreover, we address the setting where 
(some) ground-truth evidence spans have been annotated.
We present methods to jointly model
the text classification and evidence sequence labeling tasks.
We show that conditioning the evidence extraction on the predicted label,
in a \textit{classify then extract} framework,
leads to improved performance over baselines 
(with as few as a hundred annotations).
Our methods also allow generating evidence for each label, 
which can be especially useful when the input exhibits 
characteristics of multiple classes, 
or when models are less certain about their predictions.

\section*{Acknowledgements}

The authors are grateful to Anant Subramanian, Alankar Jain,
Mansi Gupta, Kundan Krishna and Suraj Tripathi 
for their insightful comments and suggestions. 
We thank Bhargavi Paranjape for her help 
in facilitating comparisons with the Information Bottleneck approach. 
Lastly, we acknowledge Highmark Health and the PwC Center
for their generous support of this line of research.

\bibliography{emnlp2020}
\bibliographystyle{acl_natbib}

\end{document}